\documentclass{article}
\usepackage{spconf,amsmath,amssymb,graphicx,booktabs}
\usepackage[T1]{fontenc}
\usepackage{microtype}
\usepackage{xcolor}
\usepackage{tikz}
\usepackage{booktabs}
\usetikzlibrary{arrows.meta,positioning,calc}
\usepackage[hidelinks]{hyperref}
\newcommand{\Rrep}{R_{\mathrm{rep}}}
\newcommand{\Lmean}{\mathcal{L}_{\mathrm{mean}}}
\newcommand{\Lrun}{\mathcal{L}_{\mathrm{run}}}

\title{Beyond Feature Reliability: Repeat-Informed Multifractal Curve Regression for Brain-Age Prediction}
\name{
    \begin{tabular}[t]{c}
        Yu Chang$^{1}$, Anzhe Cheng$^{1}$, Jiahao Chen$^{2}$, Heng Ping$^{1}$, Peiyu Zhang$^{1}$, Puquan Pan$^{1}$,\\ Tamoghna Chattopadhyay$^{1}$, Sophia Thomopoulos$^{1}$, Shahin Nazarian$^{1}$, Paul Thompson$^{1}$, Paul Bogdan$^{1}$
    \end{tabular}
    }
\address{
$^{1}$University of Southern California, USA\\
$^{2}$University of Toronto, Canada
}

\begin{document}
\maketitle

\begin{abstract}

Brain-age prediction from resting-state fMRI provides a quantitative framework for characterizing age-related changes in spontaneous brain dynamics and for identifying functional signatures. Existing studies have linked fractal and multifractal scaling to age and examined the reliability of individual features. However, prediction repeatability depends on how features fluctuate jointly and how a predictor combines them, which feature-wise reliability assessments do not capture. 
To address this problem, we propose \textbf{R}epeat-informed \textbf{M}ultifractal \textbf{C}urve \textbf{R}egression (RMCR), a structured framework for learning stable age-predictive patterns from multifractal curves. By jointly modeling curve structure and repeat-scan variability, RMCR learns predictive combinations of fluctuation orders that target both accuracy and within-subject consistency.
Relative to a matched run-level ridge baseline, RMCR reduces single-run MAE by 6.1\% on HCP-A and 7.9\% on an external Cam-CAN cohort, and within-visit repeat absolute difference by 18.5\% on HCP-A, using a single scan at inference.

\end{abstract}


\vspace{-3mm}
\section{Introduction}
\label{sec:introduction}
\vspace{-2mm}


Predicting brain age from resting-state functional magnetic resonance imaging (rs-fMRI), which captures spontaneous brain activity noninvasively, offers a quantitative approach to characterizing age-related differences in brain function.
Functional-connectivity models have predicted adult brain age~\cite{monti2020interpretable}, and the temporal organization of spontaneous fluctuations provides further candidate features~\cite{omidvarnia2024individual}. However, a useful predictor must separate age-related differences between individuals from scan-to-scan variability within the same individual: chronological age is effectively unchanged within a visit, yet variations in the measured dynamics can yield inconsistent predictions. Learning temporal representations that support both accurate and repeatable age prediction is therefore critical.


Multifractal analysis describes temporal structure beyond a single scaling exponent by characterizing how fluctuations of different magnitudes behave across temporal scales. In multifractal detrended fluctuation analysis (MF-DFA), this structure is represented by a generalized Hurst curve $h(q)$, where the statistical order $q$ controls the emphasis on smaller or larger local fluctuations~\cite{kantelhardt2002multifractal}. Age-related differences in multifractal properties~\cite{suckling2008endogenous}, together with complementary information from detrended fluctuation analysis and spectral features~\cite{cauzzo2026dfa}, motivate using these curves for brain-age prediction; their value, however, depends on learning which aspects carry age information consistently across scans.


Existing feature extraction and regression approaches face two challenges. First, extracting a multifractal curve does not ensure that the predictor uses its structure effectively. The second-order exponent $h(2)$ provides a reference scaling level, whereas variation across other orders describes the relative scaling of smaller and larger fluctuations. A scalar summary can discard informative shape differences, while ordinary ridge regression on all sampled orders keeps this information but ignores the ordering of $q$: it limits coefficient magnitude but does not explicitly discourage unstable weight changes between neighboring orders. Second, feature-wise reliability~\cite{noble2019decade} does not determine prediction repeatability~\cite{taxali2021boost}. Multifractal estimates are sensitive to sampling parameters and recording length, and their reliability varies with the inter-scan interval~\cite{guan2025fractal}. Because variations across orders and brain networks can reinforce or offset one another depending on the regression weights, neither assessing features separately nor shrinking coefficients uniformly directly controls the resulting prediction variability.

\begin{figure*}[t]
    \centering
    \includegraphics[width=\textwidth]{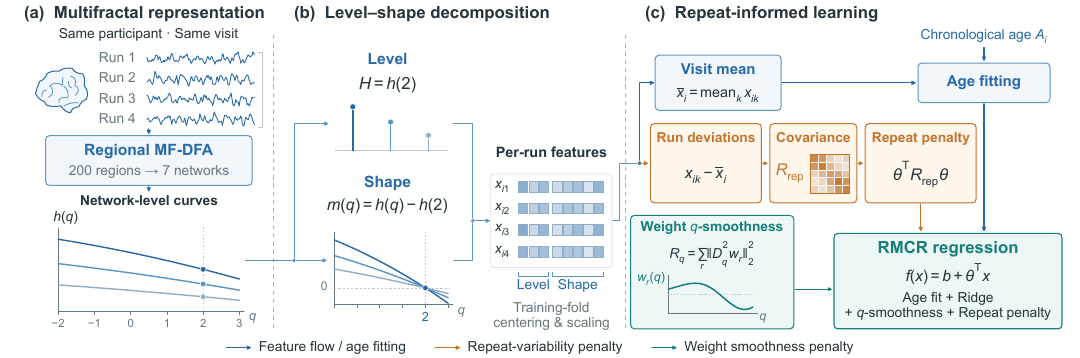}
    \caption{RMCR overview. Visit means fit chronological age; within-visit repeat deviations regularize prediction variance.}
    \label{fig:rmcr}
    \vspace{-3mm}
\end{figure*}

To address these limitations, we introduce repeat-informed multifractal curve regression (RMCR), a structured framework for learning stable age-predictive patterns from rs-fMRI. First, we formulate multifractal brain-age prediction as structured curve regression: RMCR separates second-order scaling from curve shape and learns smoothly varying weights across fluctuation orders, discouraging irregular coefficient patterns among closely related orders. Second, unlike post hoc reliability assessments~\cite{noble2019decade,taxali2021boost}, RMCR uses repeated scans during training to estimate joint feature variability across orders and brain networks, and we derive an adjustable repeat-informed penalty equal to the mean within-visit prediction variance, whose weight run-level ridge implicitly fixes to one. The resulting regularizer discourages predictive combinations that amplify scan-to-scan fluctuations. RMCR needs repeated scans only for training and predicts age from a single scan. Experiments show that RMCR reduces MAE by 6.1\% on HCP-A and 7.9\% on Cam-CAN, and lowers within-visit repeat absolute difference (RAD) by 18.5\% on HCP-A, compared with a matched run-level ridge baseline (Run Ridge).



\section{Method}
\label{sec:method}
\vspace{-0.5mm}

\subsection{MF-DFA Curves and Level--Shape Coordinates}
For each run, MF-DFA integrates a demeaned regional time series and fits a
linear trend in each of $2N_s$ forward/backward segments, where
$N_s=\lfloor L/s\rfloor$ for $L$ time points and scale $s$.
Let $v_\nu(s)$ be the mean squared detrended residual in segment $\nu$.
The fluctuation functions are~\cite{kantelhardt2002multifractal}
\begin{equation}
\begin{aligned}
F_q(s)&=\left[\frac{1}{2N_s}\sum_{\nu=1}^{2N_s}
v_\nu(s)^{q/2}\right]^{1/q}, &&q\ne0,\\
\log F_0(s)&=\frac{1}{4N_s}\sum_{\nu=1}^{2N_s}\log v_\nu(s).
\end{aligned}
\label{eq:mfdfa}
\end{equation}
Ordinary least squares of $\log F_q(s)$ against $\log(s\,\mathrm{TR})$
estimates $h(q)$ for $\mathcal Q=\{-2,-1,0,1,2,3\}$.

The scale grid contains 12 log-spaced durations from 16 to 96~s, mapped to distinct nearest-integer scales using each run's TR. Scales require $s \ge 8$ and $N_s \ge 4$; a run requires at least eight valid scales. Residual variances are floored at $10^{-12}$ to protect negative orders and logarithms. A run is excluded if any regional series is constant or any regional fit reaches this floor or is nonfinite, or if its framewise displacement (FD)~\cite{power2012spurious} exceeds 0.2~mm on average or 0.5~mm in any frame. Retained runs are analyzed on their original sampling grid without frame censoring, interpolation, band-pass filtering, or global signal regression, each of which can alter scaling within the analyzed range. MF-DFA curves are estimated in 200 Schaefer regions and averaged within seven networks~\cite{schaefer2018localglobal}.

For participant $i$, run $k$, and network $r$, define
\begin{equation}
H_{ikr}=h_{ikr}(2),\qquad m_{ikr}(q)=h_{ikr}(q)-h_{ikr}(2).
\label{eq:levelshape}
\end{equation}
By construction, shape is invariant to adding a constant to $h(q)$.
Training-fold means center all coordinates. Each level has unit variance;
the five nonzero shape coordinates share a network-specific scale equal
to the root mean of their training variances. Consequently $m(2)=0$
remains a zero feature. Below, $x_{ik}$ stacks these normalized coordinates.

\subsection{Structured Regression and Repeat Weighting}
The predictor is $f(x)=b+\theta^\top x$, with level coefficients $a_r$
and six shape coefficients $w_r$. We impose
\begin{equation}
\mathbf1^\top w_r=0,\qquad w_r=B\gamma_r,\qquad
B^\top B=I_5,\quad \mathbf1^\top B=0.
\label{eq:basis}
\end{equation}
$B$ is obtained by thin QR factorization of $[I_5;-\mathbf1^\top]$.
The zero-sum constraint selects a unique coefficient representation and makes the shape term a contrast of the curve itself ($\mathbf{1}^\top w_r = 0$ implies $w_r^\top m_r = w_r^\top h_r$), with $w_r(2)$ as its weight on $h(2)$.
We penalize $\mathcal R_q=\sum_r\|D_q^2w_r\|_2^2$, where $D_q^2$
is the second-difference operator on the ordered $q$ grid~\cite{goldsmith2011penalized}.

Let $\bar x_i=K_i^{-1}\sum_kx_{ik}$. With $N$ training participants,
each having repeated runs, define
\begin{equation}
\Rrep=\frac1N\sum_i\frac1{K_i}\sum_k
(x_{ik}-\bar x_i)(x_{ik}-\bar x_i)^\top.
\label{eq:rrep}
\end{equation}
Its quadratic form is exactly the mean within-visit prediction variance:
$\theta^\top\Rrep\theta=N^{-1}\sum_i K_i^{-1}
\sum_k[f(x_{ik})-f(\bar x_i)]^2$.
RMCR minimizes
\begin{equation}
\begin{split}
\Lmean+\lambda_0\|\theta\|_2^2+\lambda_q\mathcal R_q
+\lambda_{\mathrm{rep}}\theta^\top\Rrep\theta,\\
\Lmean=\frac1N\sum_i[A_i-f(\bar x_i)]^2,
\end{split}
\label{eq:objective}
\end{equation}
where $A_i$ is chronological age, $\lambda_0>0$, and
$\lambda_q,\lambda_{\mathrm{rep}}\ge0$.

For the same participant weights and linear predictor,
\begin{equation}
\underbrace{\frac1N\sum_i\frac1{K_i}\sum_k
[A_i-f(x_{ik})]^2}_{\Lrun}
=\Lmean+\theta^\top\Rrep\theta.
\label{eq:run_identity}
\end{equation}
Thus, matched Run Ridge is exactly the case $(\lambda_q,\lambda_{\mathrm{rep}})=(0,1)$:
run-level fitting, like training with input noise~\cite{bishop1995noise}, already
imposes a conditional-variance penalty~\cite{heinze2021core} with its weight fixed
to one. RMCR frees this weight, selecting it by inner validation, and additionally
regularizes the ordered coefficient curve.

Substituting $\theta=P\eta$, where the orthonormal block map $P$ keeps level coefficients and maps each $\gamma_r$ through $B$, turns (\ref{eq:objective}) into an unconstrained quadratic problem in $\eta$ with a closed-form solution; the intercept is unpenalized.
\vspace{-0.7mm}
\section{Experiments}
\label{sec:experiments}

\subsection{Experimental Setup}
 We use 480 HCP-Aging (HCP-A) participants with four same-visit runs (two sessions, each with AP and PA phase encoding) for development and 320 participants from the Cam-CAN repository with one run for external testing, restricted to ages 36--85 years~\cite{harms2018lifespan,bookheimer2019lifespan,taylor2017camcan, shafto2014camcan}. Of the 725 HCP-Aging Lifespan 2.0 participants (ages 36--100+), 603 were aged 36--85 with four resting-state runs, and 480 passed all checks in Sec.~2.1 (age $58.1\pm11.0$ years; mean FD $0.12\pm0.03$~mm). Of 452 Cam-CAN participants aged 36--85 with resting-state data, 320 passed the same checks (age $60.6\pm11.5$ years).
 HCP-A runs (3T Prisma, TR 0.8~s, 488 frames) are minimally preprocessed with the HCP pipelines~\cite{glasser2013minimal} and denoised with ICA-FIX~\cite{salimi2014fix} on the fs\_LR surface; Cam-CAN runs (3T TIM Trio, TR 1.97~s, 261 frames) are preprocessed with fMRIPrep~\cite{esteban2019fmriprep} in MNI space, followed by
regression of 24 motion parameters and mean white-matter and CSF signals. Because MF-DFA scales are defined in seconds (Sec.~2.1), the external test changes the scanner, TR, number of frames, and preprocessing pipeline but not the analyzed 16--96 s range, which matters because multifractal estimates depend on sampling parameters~\cite{guan2025fractal}.

\begin{figure}[t]
\centering
\includegraphics[width=0.9\columnwidth]{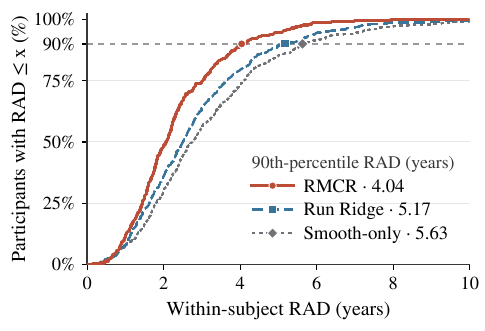}
\vspace{-5mm}
\caption{Within-visit RAD distributions. Each participant contributes the mean absolute difference over six pairs of four same-visit runs.}
\label{fig:repeatability}
\vspace{-5mm}
\end{figure}

Nested cross-validation uses five outer and three inner
participant-grouped folds. All methods share folds,
participant weights, and available runs; feature learning,
scaling, and covariance estimation use training participants
only. Inner-validation single-run MAE selects hyperparameters.
The search grids are $\lambda_0\in\{10^{-4},\ldots,10^3\}$,
$\lambda_q\in\{0,10^{-4},\ldots,10^3\}$, and
$\lambda_{\mathrm{rep}}\in\{0,\tfrac14,\tfrac12,1,2,4,8,16\}$,
with integer powers of ten. Cam-CAN predictions average the five outer-fold HCP-A models; Cam-CAN is used only for scoring, without age-bias correction.

Our feature-based ridge baselines use participant-weighted run-level loss. Level features are $h(2)$; the shape summaries are $h(-2)-h(3)$ and $h(0)-h(2)$. The spectral baseline adds network-averaged log band power and log--log spectral slope over 0.01--0.08~Hz (Welch, 128 s windows). Multiscale entropy (MSE) computes sample entropy ($m=2$, $r=0.15$ SD) at 2--10~s coarse-graining durations, averaged within networks. Functional connectivity (FC) uses the 19,900 Fisher-$z$ transformed correlations between the 200 regions.

Published-method adaptations include weighted permutation entropy (wPE) + Ridge and modular hierarchical analysis (MHA) + Linear. The former uses normalized wPE in each region, with order four and a one-frame delay, followed by ridge regression~\cite{omidvarnia2024individual}. MHA learns a shared latent-network basis from training runs and predicts age from run-specific network-activity features using participant-weighted linear regression~\cite{monti2020interpretable}. Its latent dimension is selected from $\{5, 10, 20, 40\}$ by inner-validation MAE. Both methods use the same regional time series and participant-grouped evaluation as RMCR.

\begin{table}[t]
\centering
\caption{Single-run prediction. MAE and RAD in years; Level: $h(2)$. Bold: best among learned models. $^\dagger$: paired bootstrap 95\% CI of the MAE difference from RMCR excludes zero; for RAD, this holds for all learned baselines.}
\label{tab:main_results}
\small
\setlength{\tabcolsep}{2pt}
\renewcommand{\arraystretch}{1.06}
\begin{tabular*}{\columnwidth}{@{\extracolsep{\fill}}lccccc@{}}
\toprule
& \multicolumn{3}{c}{HCP-A ($n=480$)}
& \multicolumn{2}{c}{Cam-CAN ($n=320$)}\\
\cmidrule(lr){2-4}\cmidrule(lr){5-6}
Method & MAE$\downarrow$ & $R^2\uparrow$ & RAD$\downarrow$
& MAE$\downarrow$ & $R^2\uparrow$\\
\midrule
Mean age & 9.19 & .000 & 0.00 & 9.81 & $-.049$\\
Level & 6.34\rlap{$^\dagger$} & .486 & 2.76 & 8.12\rlap{$^\dagger$} & .185\\
Level + summaries & 5.93\rlap{$^\dagger$} & .554 & 2.92 & 7.56\rlap{$^\dagger$} & .307\\
Level + spectral & 5.71\rlap{$^\dagger$} & .583 & 2.79 & 7.24\rlap{$^\dagger$} & .346\\
MSE + Ridge & 5.84\rlap{$^\dagger$} & .559 & 3.06 & 7.43\rlap{$^\dagger$} & .339\\
FC + Ridge & 5.51 & .614 & 3.10 & 6.90 & .419\\
wPE + Ridge~\cite{omidvarnia2024individual}
& 5.62\rlap{$^\dagger$} & .598 & 2.98 & 7.15 & .367\\
MHA + Linear~\cite{monti2020interpretable}
& 5.39 & .632 & 2.80 & 6.88 & .424\\
Run Ridge & 5.43\rlap{$^\dagger$} & .620 & 2.86 & 7.06\rlap{$^\dagger$} & .372\\
\midrule
\textbf{RMCR} & \textbf{5.10} & \textbf{.663} & \textbf{2.33}
& \textbf{6.50} & \textbf{.481}\\
\bottomrule
\end{tabular*}
\vspace{-5mm}
\end{table}

MAE averages absolute single-run errors within participants
and then across participants; $R^2$ uses the same weights.
Repeat absolute difference is
\begin{equation}
\mathrm{RAD}_i=\frac{2}{K_i(K_i-1)}
\sum_{k<\ell}|\hat A_{ik}-\hat A_{i\ell}|.
\label{eq:rad}
\end{equation}
Mean RAD measures within-visit agreement. Method differences are reported with 95\% confidence intervals (CIs) from a paired bootstrap that resamples participants, with all their runs and every method's predictions, 10,000 times; Fig.~3 uses the same procedure.
\subsection{Accuracy and Within-Visit Agreement}

Table~\ref{tab:main_results} compares single-run prediction across cohorts. RMCR achieves MAEs of 5.10 years on HCP-A and 6.50 years on Cam-CAN, reducing the matched Run Ridge errors by 0.33 years (95\% CI [0.14, 0.52]; 6.1\%) and 0.56 years ([0.18, 0.94]; 7.9\%), respectively, and has the lower MAE in all five HCP-A outer folds. MHA + Linear has the lowest MAE among the non-RMCR methods in both cohorts (5.39 and 6.88 years); RMCR's MAEs are lower by 0.29 and 0.38 years, respectively.

On HCP-A, RMCR also reduces mean RAD from 2.86 to 2.33 years, an improvement of 0.53 years (95\% CI [0.42, 0.64]; 18.5\%) over Run Ridge and of 0.47 years ([0.33, 0.61]) over MHA + Linear. FC + Ridge, among the most accurate baselines, has the highest RAD (3.10 years), consistent with the limited test--retest reliability of connectivity features~\cite{noble2019decade}. RMCR attains lower MAE and RAD with only 42 coefficients (7 networks $\times$ 6) versus 19,900 FC features. Figure~\ref{fig:repeatability} shows that the reduction extends to participants
with larger between-run prediction differences: RMCR lowers the 90th-percentile
RAD from 5.17 years (Run Ridge) to 4.04 years (Smooth-only: 5.63).
With a steeper prediction--age slope than Run Ridge (0.67 vs.\ 0.62; Table~2), RMCR achieves this lower RAD with less compression toward the mean age, even though its hyperparameters are selected on MAE alone. Its brain-age gap is also unrelated to mean FD after adjusting for age (partial $r=0.04$, $p=0.38$).

\subsection{Component and Representation Comparisons}
\begin{table}[t]
\centering
\caption{HCP-A component comparisons (identical inputs and constraints). $\star$: inner-validation selection ($\lambda_0$ always selected). MAE and RAD in years; Slope: visit-mean predicted age on chronological age. Diagonal keeps only the diagonal of $R_{\mathrm{rep}}$. All MAE and RAD differences from RMCR have paired bootstrap 95\% CIs excluding zero.}
\label{tab:component_ablation}
\small
\setlength{\tabcolsep}{2.5pt}
\renewcommand{\arraystretch}{1.07}
\begin{tabular*}{\columnwidth}{@{\extracolsep{\fill}}lccccc@{}}
\toprule
Variant & $\lambda_q$ & $\lambda_{\mathrm{rep}}$ & MAE$\downarrow$ & RAD$\downarrow$ & Slope\\
\midrule
Mean Ridge & 0 & 0 & 5.87 & 3.65 & 0.56\\
Run Ridge & 0 & 1 & 5.43 & 2.86 & 0.62\\
Smooth-only & $\star$ & 0 & 5.56 & 3.20 & 0.60\\
Smooth Run Ridge & $\star$ & 1 & 5.38 & 2.68 & 0.63\\
Repeat-only & 0 & $\star$ & 5.42 & 2.61 & 0.62\\
Diagonal & $\star$ & $\star$ & 5.40 & 2.55 & 0.63\\
\midrule
\textbf{RMCR} & $\star$ & $\star$ & \textbf{5.10} & \textbf{2.33}
& \textbf{0.67}\\
\bottomrule
\end{tabular*}
\end{table}

Table~\ref{tab:component_ablation} separates the contributions
of curve smoothing and repeat weighting. Smooth Run Ridge
combines smoothing with ordinary run-level fitting, fixing
$\lambda_{\mathrm{rep}}=1$. Relative to this matched comparator, RMCR reduces MAE from 5.38 to 5.10 years (by 0.28 years, 95\% CI [0.13, 0.43]; 5.2\%) and RAD from 2.68 to 2.33 years (by 0.35 years, [0.27, 0.43]; 13.1\%); this comparison isolates the benefit of
adjustable repeat weighting beyond smoothing. Inner validation selected $\lambda_{\mathrm{rep}}=8$ in three outer folds and 4 in the other two, i.e., 4--8 times the weight fixed by run-level fitting.
Without smoothing, adjustable repeat weighting mainly lowers RAD (Repeat-only: 2.61 vs.\ 2.86 years) while leaving MAE nearly unchanged (5.42 vs.\ 5.43); combined with smoothing, it lowers both, so the two penalties are complementary: smoothing restricts shape weights to smooth contrasts, and the repeat penalty selects the stable ones among them. Relative to Diagonal, RMCR reduces MAE by 0.30 years and RAD by 0.22 years, so modeling the full repeat covariance contributes to both gains.

\begin{figure}[t]
\centering
\includegraphics[width=0.9\columnwidth]{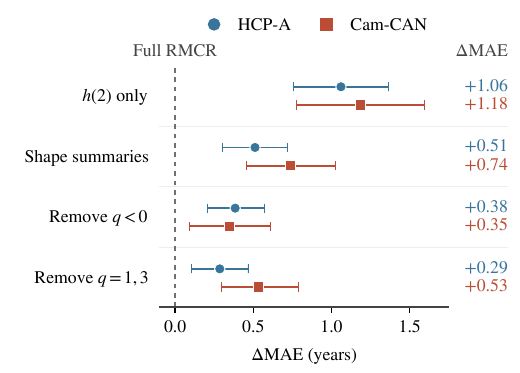}
\vspace{-3mm}
\caption{Representation comparisons: $\Delta$MAE (reduced-input minus full-RMCR MAE) with 95\% paired participant-level bootstrap CIs.}
\label{fig:representation_ablation}
\vspace{-1mm}
\end{figure}

Figure~\ref{fig:representation_ablation} compares reduced inputs within RMCR. The full curve achieves a lower MAE than every reduced input in both cohorts, with all CIs above zero; its advantage is largest over $h(2)$ alone (1.06/1.18 years on HCP-A/Cam-CAN) and persists against the variants that drop the negative orders or $q = 1, 3$, so the curve carries age information beyond both $h(2)$ and its summaries.

\vspace{-2mm}
\section{Conclusion}
\label{sec:conclusion}
We presented RMCR, a brain-age regression framework combining
network-level $h(2)$ features with multifractal curve shape.
Smoothness across fluctuation orders and a covariance-based repeat
penalty jointly lower MAE in both cohorts and within-visit
RAD on HCP-A relative to run-level ridge.

\clearpage
\section{Compliance with Ethical Standards}
This retrospective study used de-identified data from the HCP-Aging~\cite{harms2018lifespan,bookheimer2019lifespan} and Cam-CAN~\cite{taylor2017camcan,shafto2014camcan} repositories under their data use agreements; the original studies obtained ethical approval and informed consent.
\bibliographystyle{IEEEbib}
\bibliography{references}
\end{document}